\pdfoutput=1

\documentclass[11pt]{article}

\usepackage[]{ACL2023}

\usepackage{times}
\usepackage{latexsym}

\usepackage[T1]{fontenc}

\usepackage[utf8]{inputenc}

\usepackage{microtype}

\usepackage{inconsolata}
\usepackage{tree-dvips}
\usepackage{qtree}
\usepackage{tikz}
\usetikzlibrary{trees}
\usepackage{CJKutf8}
\usepackage[linguistics]{forest}
\usepackage{enumitem}
\usepackage{tikz}
\usepackage{fontawesome5}
\usepackage{xcolor}
\usepackage{color} 
\usepackage{soul} 
\usepackage{booktabs,colortbl}
\usepackage{times}
\usepackage{latexsym}
\usepackage{microtype}
\usepackage{amsmath}
\usepackage{amssymb}
\usepackage{multirow}
\usepackage{multicol}
\usepackage{bbding}
\usepackage{graphicx}
\usepackage{xfrac}
\usepackage{subfigure}
\usepackage{makecell}
\usepackage{cleveref}
\usepackage{arydshln}
\usepackage{color}
\usepackage{booktabs}
\usepackage{cleveref}
\usepackage{tabu}
\usepackage{tabularx}
\usepackage{booktabs}
\usepackage{float}
\newcommand{\dashrule}[1][black]{%
  \color{#1}\rule[\dimexpr.5ex-.2pt]{4pt}{.4pt}\xleaders\hbox{\rule{4pt}{0pt}\rule[\dimexpr.5ex-.2pt]{4pt}{.4pt}}\hfill\kern0pt%
}
\title{Exploring the Use of Large Language Models for Reference-Free Text Quality Evaluation: An Empirical Study}

\author{Yi Chen$^{\heartsuit\spadesuit}$\thanks{\ \ Equal Contribution.}, Rui Wang$^{\heartsuit\spadesuit *}$, Haiyun Jiang\thanks{\ \ Corresponding Authors.}, Shuming Shi, Ruifeng Xu$^{\heartsuit\clubsuit\spadesuit \dagger}$\\
  $^{\heartsuit}$Harbin Institute of Technology, Shenzhen, China\\
  $^{\clubsuit}$Peng Cheng Laboratory, Shenzhen, China\\
  $^{\spadesuit}$Guangdong Provincial Key Laboratory of Novel Security Intelligence Technologies\\
  {\tt yichennlp@gmail.com,ruiwangnlp@outlook.com,xuruifeng@hit.edu.cn}\\
 }

\begin{document}
\maketitle
\begin{abstract}

Evaluating the quality of generated text is a challenging task in NLP, 
due to the inherent complexity and diversity of text. 
Recently, large language models (LLMs) have garnered significant attention due to their impressive performance in various tasks.
Therefore, we present this paper to investigate the effectiveness of LLMs, especially ChatGPT, and explore ways to optimize their use in assessing text quality.
We compared three kinds of reference-free evaluation methods.
The experimental results prove that ChatGPT is capable of evaluating text quality effectively from various perspectives without reference and 
demonstrates superior performance than most existing automatic metrics. 
In particular, the Explicit Score, which utilizes ChatGPT to generate a numeric score measuring text quality, is the most effective and reliable method among the three exploited approaches. 
However, directly comparing the quality of two texts may lead to suboptimal results.
We believe this paper will provide valuable insights
for evaluating text quality with LLMs and have released the used data\footnote{\url{https://github.com/MilkWhite/LLMs\_for\_Reference\_Free\_Text\_Quality\_Evaluation}}.
\end{abstract}

\section{Introduction}
Automated evaluation of text generation quality has posed a long-standing challenge in the field of natural language processing (NLP). 
On the one hand, the diverse forms of textual expression make it impossible for reference-based methods to account for all possible situations\cite{bert-score, NEURIPS2021_e4d2b6e6, chen2022mcpg}. 
On the other hand, devising reliable metrics without reference is not a straightforward task and can also be problematic \cite{sun-zhou-2012-joint, niu-etal-2021-unsupervised, shen-etal-2022-evaluation}. 
Furthermore, different types of text necessitate evaluation of distinct aspects, e.g. coherence, fluency, and consistency \cite{fabbri-etal-2021-summeval, mehri-eskenazi-2020-unsupervised, Wang2023RetrievalfreeKI}, which makes it hard to design metrics for each type of text and dimension separately.

Nowadays, large language models (LLMs) \cite{brown2020language, ouyang2022training, chung2022scaling, chowdhery2022palm, zhang2022opt, touvron2023llama, du-etal-2022-glm} represented by  ChatGPT\footnote{\url{https://openai.com/blog/chatgpt}} have revolutionized the field of NLP by achieving remarkable results in a wide range of NLP tasks \cite{song2023chatgpt,chen2022would}. 
Recent studies \cite{fu2023gptscore, wang2023chatgpt, kocmi2023large, ji2023exploring} have also demonstrated the potential of LLMs in evaluating the quality of generated texts. 
In this paper, we present an empirical study that compares different methods for text quality evaluation using LLMs 
in a reference-free mode.
The key insights from our empirical findings are as follows:

\begin{itemize}[wide=0\parindent]

    \item \textbf{How accurately can ChatGPT assess text quality without references? } (\S \ref{sec:experiment_indicidual_score})
    
    It is feasible for ChatGPT to evaluate text quality without reference, and it outperforms commonly used metrics even with a simple prompt design.

    
    \item \textbf{What is the most effective approach to evaluate text quality using ChatGPT?} (\S \ref{sec:main_experiments})
    

    Generally, using ChatGPT to generate an explicit score for text quality is the best and most stable method among the three we compared. We suggest using greedy decoding for more reliable results.
    

    \item \textbf{Why may directly comparing two texts using ChatGPT yield suboptimal results?} (\S \ref{sec:analyais_pairwise_comparison})


    Due to its strict standard for ``high-quality'' text, ChatGPT often considers most generated texts unsatisfactory. Therefore, distinguishing between two subpar texts becomes challenging for ChatGPT.
    

    \item \textbf{Why is Implicit Score generally less effective than Explicit Score?} (\S \ref{sec:analysis_implicit_score})



    Compared to generating an Explicit Score with ChatGPT, using the confidence of text-davinci models to determine text quality (Implicit Score) is less effective due to different distribution characteristics. Implicit Score has a narrow range and peak structure, while Explicit Score allows better differentiation with its smoother distribution.

    \item \textbf{How can prompt design impact ChatGPT in generating an Explicit Score?} (\S \ref{sec:analysis_explicit_score})
    


    When prompting ChatGPT for an Explicit Score, it would be better to avoid detailed scoring criteria if such criteria lack clear definitions for each score range.
    A general description of the evaluation standard is enough. 
    Also, making ChatGPT provide justifications in a "chain-of-thought" manner before scoring can lead it to prioritize its reasoning process over the text. These justifications tend to be templated and similar across different texts, reducing the discriminative power of the final score.
    
\end{itemize}

\section{Method}
We explore two different reference-free paradigms, i.e., \textit{Individual Score} and \textit{Pairwise Comparison} for text evaluation using ChatGPT and text-davinci models.
Individual Score assesses the quality of a single text by a numerical score, while Pairwise Comparison focuses on the relative quality of two texts and requires a direct comparison to determine which one is superior. 
Within the Individual Score paradigm, two methods are typically exploited: \textit{Explicit Score}, obtained through direct text generation, and \textit{Implicit Score}, obtained through the token probabilities outputted by the model. 

\subsection{Individual Score}
\paragraph{Explicit Score}
Conditioned on a given input text (optional), we prompt ChatGPT to directly generate a score to measure the absolute quality of each text individually in terms of a specific aspect or the overall performance. 
An example prompt designed for scoring the overall quality of a storyline is shown as follows:
\begin{table}[!h]
\small
    \centering
    \colorbox{gray!8}{
    \begin{tabular}{@{}p{7.3cm}}
    ========= \textsc{Prompt for Explicit Score} =========\\\\
    \vspace{-2mm}
    Score the following storyline given the beginning of the story on a continual scale from 0 (worst) to 100 (best), where a score of 0 means "The storyline makes no sense and is totally not understandable" and a score of 100 means "The storyline is perfect-written and highly consistent with the given beginning of the story".

    
    
    \end{tabular}}
\end{table}

\begin{table}[!h]
\small
    \centering
    \colorbox{gray!8}{
    \begin{tabular}{@{}p{7.3cm}}

    The beginning of the story: \\
    \texttt{[Conditioned Text]}\\\\
    
    Storyline: \\
    \texttt{[Generated Text]}\\\\
    
    Score:
    \end{tabular}}
\end{table}
    
    
    
\vspace{-1.0cm}
\paragraph{Implicit Score}

Given the LLM's potential insensitivity to numerical values and the lack of explicit instructions for aligning score intervals with specific criteria, score fluctuations may occur across different samples. 
Therefore, we propose an alternative approach by framing the problem as a binary Yes or No question, where the confidence level of answering "yes" serves as the Implicit Score.
 An illustrative example is presented below:
 \vspace{-0.05cm}
\begin{table}[!htbp]
\small
    \centering
    \colorbox{gray!8}{
    \begin{tabular}{@{}p{7.3cm}}
    ========= \textsc{Prompt for Implicit Score} =========\\\\

    Consider the following storyline written according to the given beginning of the story:\\\\

    The beginning of the story: \\
    \texttt{[Conditioned Text]}\\\\
    
    Storyline: \\
    \texttt{[Generated Text]}\\\\
    
    Question: Is the storyline well-written and consistent with the beginning of the story?\\\\
    
    Answer:
    \end{tabular}}
\end{table}
\vspace{-0.05cm}
\\Unfortunately, access to ChatGPT's token probabilities is currently unavailable. Text-davinci-003 is similar to ChatGPT in that they are both trained through supervised instruction tuning and Reinforcement Learning from Human Feedback (RLHF) based on GPT-3.5, and they both exhibit excellent performance in following and fulfilling human instructions.
Therefore, we utilize text-davinci-003 to derive the Implicit Score as a baseline metric instead.
To facilitate a more comprehensive comparison, we also obtain the Implicit Score from text-davinci-001, an earlier version of the text-davinci series model which is based on GPT-3 and has not been trained using RLHF.
Due to a limitation of the OpenAI API, only the top 5 most probable tokens are returned with log probabilities. Therefore, we instead estimate the Implicit Score using the following formula:
\begin{equation}
\small
\begin{split}
     p(\text{yes}) &= \sum_{t \in {{\mathcal{T}_{top5}} \cap \mathcal{T}_{yes}}}{p(t)}, \\
     p(\text{no})  &= \sum_{t \in {{\mathcal{T}_{top5}} \cap \mathcal{T}_{no}}}{p(t)}, \\
     \text{Implicit Score} &= \max(p(yes), 1-p(no)). \\
\end{split}
\end{equation}
\\Here, $p(t)$ represents the probability of predicting token $t$ immediately following the prompt "Answer:". The sets $\mathcal{T}_{yes}$ and $\mathcal{T}_{no}$ consist of the affirmative and negative response tokens, respectively, i.e., 
$\mathcal{T}_{yes} = \{\text{`` Yes'', ``Yes'', `` yes'', ``yes''}\}$, and $\mathcal{T}_{no} = \{\text{`` No'', ``No'', `` no'', ``no''}\}$.

\subsection{Pairwise Comparison}
Another paradigm to assess text quality is by directly comparing a pair of generated texts based on the same input. This method primarily focuses on the relative quality of the texts. For instance, a prompt for comparing the overall quality of two storylines written according to the same initial story beginning is shown as follows:
\vspace{-0.1cm}
\begin{table}[!htbp]
\small
    \centering
    \colorbox{gray!8}{
    \begin{tabular}{@{}p{7.3cm}}
    ====== \textsc{Prompt for Pairwise Comparison} ======\\\\

    Consider the following two storylines written according to the given beginning of the story:\\\\

    The beginning of the story:\\
    \texttt{[Conditioned Text]}\\\\
    
    Storyline-1:\\
    \texttt{[Generated Text-1]}\\\\
    
    Storyline-2:\\
    \texttt{[Generated Text-2]}\\\\
    
    Question: Which storyline is better-written and more consistent with the beginning of the story? Please answer with one of the following options.\\\\
    
    Options:\\
    (A) Storyline-1\\
    (B) Storyline-2\\
    (C) Both storylines are equally well-written and consistent with the beginning of the story.\\\\
    
    Answer: I will choose Option
    \end{tabular}}
\end{table}
    

    
    
    
    
    

\section{Experimental Setup}
\subsection{Tasks and Datasets}
We conduct experiments on four distinct natural language generation tasks: Text Summarization, Dialogue Response Generation, Story Generation, and Paraphrase Generation.
\paragraph{Text Summarization} aims to summarize the key points of a given long text. 
SummEval \cite{fabbri2021summeval} is a collection of human annotations for 16 model-generated summaries on 100 CNN/DaliyMail news over 4 dimensions: coherence (COH), fluency (FLU), consistency (CON), and relevance (REL). 
Due to the budget limit, we randomly sample 20 news and corresponding annotations from SummEval for evaluation.


\paragraph{Dialogue Response Generation} aims to generate a response based on the preceding dialogue. 
We conduct experiments on the dialogue-level FED dataset \cite{mehri-eskenazi-2020-unsupervised}, which contains fine-grained human judgments for 124 conversations. The evaluation aspects include coherence (COH), error recovery (ERR), consistency (CON), diversity (DIV), topic depth (DEP), likeability (LIK), understanding (UND), flexibility (FLE), informativeness (INF), inquisitiveness (INQ) and overall performance (Overall). 
However, we do not include ERR in our evaluation since some annotations 
are missing.

\paragraph{Story Generation} aims to automatically write a storyline based on a given beginning of the story. 
We employ OpenMEVA-ROC \cite{guan-etal-2021-openmeva} for evaluation, which contains 200 story beginnings and 5 corresponding machine-generated storylines for each beginning. 
Each storyline is manually annotated in terms of overall quality.

\paragraph{Paraphrase Generation} aims to rephrase a sentence in different words or forms while preserving its original meaning. 
We use Twitter-Para \cite{xu-etal-2014-extracting, xu-etal-2015-semeval} for evaluation, containing 761 input sentences and each input has 9.41 paraphrase candidates on average. 
We adopt the test set \cite{shen-etal-2022-evaluation} extended from Twitter-Para by adding 20\% of the input sentences as candidates, denoted as Twitter (Extend).


\subsection{Chosen Metrics}
Following the settings of previous works, we select baseline metrics from the following widely used metrics accordingly: \textbf{ROUGE-1}, \textbf{ROUGE-2} and \textbf{ROUGE-L} \cite{lin-2004-rouge}; \textbf{BERTScore} \cite{bert-score}; \textbf{MoverScore} \cite{zhao-etal-2019-moverscore}; \textbf{PRISM} \cite{thompson-post-2020-automatic}; \textbf{BARTScore} and its enhanced versions, \textbf{BARTScore+CNN} and \textbf{BARTScore+CNN+Para} \cite{NEURIPS2021_e4d2b6e6}; \textbf{BERT-R} \cite{ghazarian-etal-2019-better}; \textbf{GPT-2} \cite{radford2019language}; \textbf{USR} \cite{mehri-eskenazi-2020-usr}; \textbf{S-DiCoh} \cite{mesgar-etal-2020-dialogue}; \textbf{FED} \cite{mehri-eskenazi-2020-unsupervised}; \textbf{DynaEval} \cite{zhang-etal-2021-dynaeval}; \textbf{SelfEval} \cite{ma-etal-2022-self}; \textbf{PPL} \cite{guan-etal-2021-openmeva}; \textbf{iBLEU} \cite{sun-zhou-2012-joint}; \textbf{BERT-iBLEU} \cite{niu-etal-2021-unsupervised}; \textbf{ParaScore} \cite{shen-etal-2022-evaluation}.
Note that, \citet{shen-etal-2022-evaluation} also use a reference-free version of BERTScore and ParaScore, denoted as \textbf{BERTScore.Free} and \textbf{ParaScore.Free}.

\subsection{Meta Evaluation}
\paragraph{Individual Score}
In order to assess the reliability of Individual Scores, we utilize the Spearman \cite{zar2005spearman} and Pearson \cite{mukaka2012guide} correlation coefficients. As SummEval and OpenMEVA provide an equivalent number of model-generated results for each input, we present the sample-level correlations for these datasets. Whereas, for Twitter (Extend) and the dialog-level FED datasets, we report the dataset-level correlations instead.

\paragraph{Pairwise Comparison} 

To avoid an excessive volume of requests when testing all permutations of pairwise comparisons in each dataset using ChatGPT, we have opted to randomly sample 200 pairs from each dataset as an approximation. 
To estimate the reliability of metrics for pairwise comparison, Kendall's Tau-b  \cite{10.2307/2332303} is employed to evaluate the correlation between two measured variables. A detailed explanation of Kendall's Tau-b is shown in Appendix \ref{app:kendall}.

\section{Main Experiments}\label{sec:main_experiments}



 	

\subsection{Individual Score}\label{sec:experiment_indicidual_score}

\begin{table}[!t]
\small
    \centering
    \begin{tabular}{l c c c c}
    \toprule
    \multirow{2}{*}{Metrics} & \multicolumn{4}{c}{Spear.} \\
    \cmidrule{2-5}
    & COH & FLU & CON & REL \\
    \midrule
    ROUGE-1 & 21.6 & 10.5 & 10.9 & \underline{{42.6}}\\
    ROUGE-2 & 30.7 & 19.1 & 20.7 & 36.9\\
    ROUGE-L & 17.4 & 10.2 & 9.6 & 40.0\\
    BERTScore & 28.5 & 10.6 & 13.4 & 29.5\\
    MoverScore & 22.5 & 11.8 & 14.6 & 39.2\\
    PRISM & 23.7 & 17.5 & 35.2 & 16.9\\
    BARTScore & 33.4 & 20.9 & 34.8 & 24.8\\
    +CNN & \underline{43.3} & \underline{28.7} & \underline{{42.7}} & 36.1\\
    +CNN+Para & {40.1} & 27.2 & 41.0 & 32.0\\
    \midrule
    \textsc{Implicit Score}\\
    [0.5ex]\cdashline{1-5}\noalign{\vskip 0.5ex}
    text-davinci-001 & -1.7 & -5.6 & 19.7 & 8.4\\
    text-davinci-003 & {\textbf{57.4}} & {\textbf{32.9}} & {35.2} & 28.1\\
    \midrule
    \textsc{Explicit Score}\\
    [0.5ex]\cdashline{1-5}\noalign{\vskip 0.5ex}
    ChatGPT (sampling) & 45.8 & 22.1 & {41.2} & {39.2}\\
    ChatGPT (greedy) & 52.2 & 19.3 & \textbf{43.3} & \textbf{46.0}\\
    \bottomrule
    \end{tabular}
    \caption{Sample-level Spearman (Spear.) correlation of different aspects on SummEval.}
    \label{tab:spearman_summeval}
\end{table}
\begin{table*}[!htbp]
\small
    \centering
    \begin{tabular}{l c c c c c c c c c c}
    \toprule
    \multirow{2}{*}{Metrics} & \multicolumn{10}{c}{Spear.}\\
    \cmidrule{2-11}
     & COH & CON & DIV & DEP & LIK & UND & FLE & INF & INQ & Overall \\
    \midrule
    BERT-R & 22.9 & 16.3 & 19.6 & 19.2 & 28.1 & 19.8 & 25.3 & 21.1 & 33.7 & 24.8\\
    GPT-2 & 12.3 & 9.1 & 14.7 & 9.7 & 17.9 & 7.0 & 13.4 & 11.6 & 7.1 & 12.3 \\
    USR & 19.4 & 16.9 & 24.2 & 34.1 & 22.1 & 17.2 & 20.9 & 28.8 & 18.8 & 28.8 \\
    S-DiCoh & 3.8 & 1.7 & 5.9 & 4.6 & -7.0 & -10.0 & 4.4 & 2.8 & -5.4 & -7.3 \\
    FED & 25.1 & 11.6 & \underline{{44.9}} & \underline{52.2} & 26.2 & 30.6 & \underline{40.8} & 33.7 & 29.8 & 44.3 \\
    DynaEval & 42.3 & \underline{35.2} & 33.2 & 43.9 & \underline{39.8} & 36.1 & 38.9 & \underline{39.6} & 38.8 & \underline{48.2} \\
    SelfEval & \underline{43.6} & 34.7 & 26.3 & 32.7 & 39.0 & \underline{40.6} & 31.7 &31.8 & \underline{42.1} & 43.5 \\
    \midrule
    \textsc{Implicit Score}\\
    [0.5ex]\cdashline{1-11}\noalign{\vskip 0.5ex}
    text-davinci-001 & 37.9 & 33.0 & 36.1 & 26.2 & 35.0 & 57.5 & 39.5 & 54.8 & 45.0 & 39.4\\
    text-davinci-003 & {46.8} & {43.8} & 24.9 & {53.4} & \textbf{57.3} & 57.6 & 45.0 & 55.1 & \textbf{59.0} & \textbf{58.0}\\
    \midrule
    \textsc{Explicit Score}\\
    [0.5ex]\cdashline{1-11}\noalign{\vskip 0.5ex}
    ChatGPT (sampling) & {57.8} & \textbf{47.8} & {44.5} & 51.5 & 47.2 & \textbf{61.7} & {49.4} & {61.7} & 42.8 & 55.8\\
     ChatGPT (greedy) & \textbf{62.4} & 47.5 & \textbf{48.3} & \textbf{55.5} & 55.4 & 60.0 & \textbf{54.8} & \textbf{62.0} & 42.3 & 54.2\\
    \bottomrule
    \end{tabular}
    \caption{Dataset-level Spearman (Spear.) correlation of different aspects on dialogue-level FED.}
    \label{tab:spearman_fed}
\end{table*}

\begin{table}[!htbp]
\small
    \centering
    \begin{tabular}{p{3cm} c c}
    \toprule
    Metrics & Spear. & Pear. \\
    \midrule
    ROUGE-1 & 1.4 & 2.0 \\
    ROUGE-2 & 3.5 & 4.1 \\
    ROUGE-L & 1.3 & 2.1 \\
    BERTScore & 14.0 & 12.0 \\
    Perplexity & \underline{32.4} & \underline{33.0} \\
    BARTScore & -6.5 & -8.2 \\
    +CNN & 4.9 & 2.6 \\
    +CNN+Para & 6.4 & 5.0 \\
    \midrule
    \textsc{Implicit Score}\\
    [0.5ex]\cdashline{1-3}\noalign{\vskip 0.5ex}
    text-davinci-001 & 30.3 & 32.9 \\
    text-davinci-003 & 37.9 & 43.4 \\
    \midrule
    \textsc{Explicit Score}\\
    [0.5ex]\cdashline{1-3}\noalign{\vskip 0.5ex}
    ChatGPT (sampling) & {47.6} & {49.0} \\
    ChatGPT (greedy) & \textbf{49.9} & \textbf{51.7} \\
    \bottomrule
    \end{tabular}
    \caption{Sample-level Spearman (Spear.) and Pearson (Pear.) correlation on OpenMEVA.}
    \label{tab:spearman_openmeva}
\end{table}

\begin{table}[!htbp]
\small
    \centering
    \begin{tabular}{p{3cm} c c}
    \toprule
    Metrics & Spear. & Pear. \\
    \midrule
    iBLEU & 3.2 & 1.1  \\
    BERTScore & 43.2 & 42.7  \\
    BERTScore.Free & 41.9 & 31.6  \\
    BARTScore+CNN+Para & 27.6 & 28.0  \\
    BERT-iBLEU & 41.6 & 32.7  \\
    ParaScore & \underline{\textbf{53.0}} & \underline{\textbf{52.7}}  \\
    ParaScore.Free & 49.5 & 49.6  \\
    \midrule
    \textsc{Implicit Score}\\
    [0.5ex]\cdashline{1-3}\noalign{\vskip 0.5ex}
    text-davinci-001 & 15.8 & 15.9 \\
    text-davinci-003 & 44.4 & 40.3 \\
    \midrule
    \textsc{Explicit Score}\\
    [0.5ex]\cdashline{1-3}\noalign{\vskip 0.5ex}
    ChatGPT (sampling) & 45.1 & 44.3 \\
    ChatGPT (greedy) & 46.5 & 45.4 \\
    \bottomrule
    \end{tabular}
    \caption{Dataset-level Spearman (Spear.) and Pearson (Pear.) correlation on Twitter (Extend).}
    \label{tab:spearman_twitter_para}
\end{table}

Notably, as shown in \Cref{tab:spearman_summeval,tab:spearman_fed,tab:spearman_openmeva,tab:spearman_twitter_para}, even without providing reference or calibration details for different score ranges, ChatGPT's Explicit Score has already correlated with human scores better than most commonly used automated metrics. 
On Twitter (Extend), it is only outperformed by ParaScore and ParaScore.Free, which requires the use of reference or hyper-parameter adjustments on a dev set. 
Additionally, the performance of the Explicit Score further improves when we use greedy search instead of Top-P sampling for decoding. 


It is worth noting that the Implicit Score based on text-davinci-003 also shows promising results. This suggests that LLMs' confidence level in determining whether a text meets a specific standard (yes or no) can reflect the text's quality to some extent. Besides, the Implicit Score based on text-davinci-003 performs better than that based on text-davinci-001 in most cases, perhaps due to RLHF, allowing text-davinci-003 to provide answers that align with human instructions better.

\subsection{Pairwise Comparison}\label{sec:experiment_pairwise_comparison}


\begin{table}[!t]
\small
    \centering
    \begin{tabular}{l c c c c}
    \toprule
    {\multirow{2}{*}{Metrics}} & \multicolumn{4}{c}{Kend.} \\
    \cmidrule{2-5}
    & COH & FLU & CON & REL \\
    \midrule
    \textsc{Implicit Score}\\
    [0.5ex]\cdashline{1-5}\noalign{\vskip 0.5ex}
    text-davinci-001 & -3.2 & -4.3 & 9.3 & 12.9\\
    text-davinci-003 & 46.9 & \textbf{24.5} & 35.3 & 29.1 \\
    \midrule
    \textsc{Explicit Score}\\
    [0.5ex]\cdashline{1-5}\noalign{\vskip 0.5ex}
    ChatGPT (sampling) & \textbf{50.3} & 8.6 & 31.7 & 44.3\\
    ChatGPT (greedy) & 43.7 & {16.8} & \textbf{32.8} & \textbf{52.5}\\
    \midrule
    \textsc{Comparison}\\
    [0.5ex]\cdashline{1-5}\noalign{\vskip 0.5ex}
    ChatGPT (sampling) & 22.6 & 7.8 & 24.2 & 30.5\\
    ChatGPT (greedy) & 34.5 & {17.4} & 22.0 & 34.0\\
    \bottomrule
    \end{tabular}
    \caption{Estimated Kendall's tau-b (Kend.) correlation of different aspects on SummEval.}
    \label{tab:pairwise_summeval}
\end{table} 

\begin{table*}[!t]
\small
    \centering
    \begin{tabular}{l c c c c c c c c c c}
    \toprule
    \multirow{2}{*}{Metrics} & \multicolumn{10}{c}{Kend.}\\
    \cmidrule{2-11}
    & COH & CON & DIV & DEP & LIK & UND & FLE & INF & INQ & Overall \\
    \midrule
    \textsc{Implicit Score}\\
    [0.5ex]\cdashline{1-11}\noalign{\vskip 0.5ex}
    text-davinci-001 & 33.3 & 32.0 & 29.6 & 25.1 & 25.6 & 49.9 & 32.8 & 44.8 & 49.5 & 33.6\\
    text-davinci-003 & 28.8 & 30.5 & 18.8 & 36.9 & 41.9 & 43.2 & 34.0 & 45.8 & 43.0 & 36.7\\
    \midrule
    \textsc{Explicit Score}\\
    [0.5ex]\cdashline{1-11}\noalign{\vskip 0.5ex}
    ChatGPT (sampling) & 48.4 & \textbf{44.1} & 32.4 & 47.5 & 46.7 & 48.0 & 36.2 & 45.6 & \textbf{45.9} & \textbf{44.2}\\
    ChatGPT (greedy) & \textbf{50.2} & 39.6 & \textbf{45.5} & \textbf{53.5} & \textbf{50.8} & \textbf{53.7} & 50.5 & \textbf{47.7} & 38.1 & 41.7\\
    \midrule
    \textsc{Comparison}\\
    [0.5ex]\cdashline{1-11}\noalign{\vskip 0.5ex}
    ChatGPT (sampling) & 28.3 & 16.1 & 28.5 & 31.5 & 43.0 & 27.5 & \textbf{55.5} & 35.2 & 24.5 & 38.6\\
    ChatGPT (greedy) & 24.3 & 13.7 & 28.5 & 33.8 & 41.9 & 27.5 & 55.5 & 34.1 & 25.6 & 37.5\\
    \bottomrule
    \end{tabular}
    \caption{Estimated Kendall's tau-b (Kend.) correlation of different aspects on dialogue-level FED.}
    \label{tab:pairwise_fed}
\end{table*}

\begin{table}[!t]
\small
    \centering
    \begin{tabular}{l c c c c}
    \toprule
    \multirow{2}{*}{Metrics} & \multicolumn{4}{c}{Kend.} \\
    \cmidrule{2-5}
    & Hard & Medium & Easy & All \\
    \midrule
    \textsc{Implicit Score}\\
    [0.5ex]\cdashline{1-5}\noalign{\vskip 0.5ex}
    text-davinci-001 & 6.3 & 29.8 & 44.4 & 16.6 \\
    text-davinci-003 & \textbf{27.9} & 36.8 & 66.7 & 33.2 \\
    \midrule
    \textsc{Explicit Score}\\
    [0.5ex]\cdashline{1-5}\noalign{\vskip 0.5ex}
    ChatGPT (sampling) & 18.5 & 47.3 & 74.3 & 31.2 \\
    ChatGPT (greedy) & 16.8 & \textbf{62.6} & \textbf{82.5} & \textbf{36.2} \\
    \midrule
    \textsc{Comparison}\\
    [0.5ex]\cdashline{1-5}\noalign{\vskip 0.5ex}
    ChatGPT (sampling) & 8.1 & 22.8 & 33.3 & 14.5 \\
    ChatGPT (greedy) & 9.9 & 29.8 & 55.6 & 19.7\\
    \bottomrule
    \end{tabular}
    \caption{Estimated Kendall's tau-b (Kend.) correlation on OpenMEVA.}
    \label{tab:pairwise_openmeva}
\end{table}

\begin{table}[!t]
\small
    \centering
    \begin{tabular}{l c c c c}
    \toprule
    \multirow{2}{*}{Metrics} & \multicolumn{4}{c}{Kend.} \\
    \cmidrule{2-5}
    & Hard & Medium & Easy & All \\
    \midrule
    \textsc{Implicit Score}\\
    [0.5ex]\cdashline{1-5}\noalign{\vskip 0.5ex}
    text-davinci-001 & 21.6 & 34.6 & 13.6 & 20.4\\
    text-davinci-003 & 25.5 & 19.2 & 59.1 & 28.6\\
    \midrule
    \textsc{Explicit Score}\\
    [0.5ex]\cdashline{1-5}\noalign{\vskip 0.5ex}
    ChatGPT (sampling) & \textbf{27.8} & \textbf{40.0} & 53.8 & \textbf{34.9} \\
    ChatGPT (greedy) & 15.3 & 38.5 & 57.0 & 31.2\\
    \midrule
    \textsc{Comparison}\\
    [0.5ex]\cdashline{1-5}\noalign{\vskip 0.5ex}
    ChatGPT (sampling) & 14.6 & 31.0 & \textbf{68.3} & 31.3\\
    ChatGPT (greedy) & 10.0 & 22.2 & 65.1 & 26.3\\
    \bottomrule
    \end{tabular}
    \caption{Estimated Kendall's tau-b (Kend.) correlation on Twitter (Extend).}
    \label{tab:pairwise_twitter_para}
\end{table}

Scoring individual samples without providing detailed criteria for each score range may lead to inconsistent evaluation standards across different samples. 
Alternatively, we hypothesize that a direct comparison of quality between a pair of samples is more likely to yield reliable evaluation results from ChatGPT. However, our analysis in \Cref{tab:pairwise_summeval,tab:pairwise_fed,tab:pairwise_openmeva,tab:pairwise_twitter_para} suggests that direct pairwise comparison is not as effective as expected, and eliminating the influence of sampling in decoding is not always advantageous for comparison.


We further categorize the texts for comparison into three levels of difficulty,
namely hard, medium, and easy, based on the difference in human scores. The larger the score difference between a pair of texts, the easier it is to discern the better one. 
The performance of various metrics on distinct difficulty levels is shown in Tables \ref{tab:pairwise_openmeva} and \ref{tab:pairwise_twitter_para}. Overall, the metrics exhibit an increasing trend in performance as the difficulty decreases.

Moreover, our investigation indicates that the Implicit Score derived from text-davinci-003 outperforms or performs comparably to the Explicit Score based on ChatGPT when comparing hard text pairs. This finding may be attributed to the higher precision of the Implicit Score, which is based on the model's output token probability (a floating-point number), as opposed to the model's generated Explicit Score, which is limited to integer values ranging from 0 to 100.



\section{Detailed Analysis}
\subsection{Why does the pairwise comparison paradigm perform worse?}\label{sec:analyais_pairwise_comparison}

\begin{table*}[!t]
\small
    \centering
    \begin{tabular}{c p{13cm}}
    \toprule
    \multicolumn{2}{c}{{\textsc{Prompts for Pairwise Comparison on Story Generation}}} \\
    \midrule
    \textsc{Prompt V1} & The default prompt where we first provide the beginning of the story and the corresponding two storylines for comparison before presenting the question. \\
    \textsc{Prompt V2} & A revised version of Prompt V1 where we first propose the question, then provide the beginning of the story and present the two storylines to be compared in the form of options. \\
    \textsc{Prompt V3} & A mirrored version of Prompt V1 where we instruct the model to choose ``which one is worse'' instead of ``which one is better'' from the two given storylines. \\
    \textsc{Prompt V4} & A ``chain-of-thought'' version of Prompt V1 where we require the model to illustrate the reasoning process before presenting the final answer.\\

    \toprule
    \multicolumn{2}{c}{\textsc{Prompts for Explicit Score on Story Generation}}\\
    \midrule
    \textsc{Prompt V1} & The default prompt where we only specify the rating criteria for zero and full marks. \\
    \textsc{Prompt V2} & A rephrased version of Prompt V1.\\
    \textsc{Prompt V3} & A simplified version of Prompt V1 where we only describe the dimensions that need to be evaluated. \\
    \textsc{Prompt V4} & A detailed prompt where we divide the scores into 5 scales and list the corresponding evaluation criteria for each score scale.\\
    \textsc{Prompt V5}
    & A ``chain-of-thought'' version of Prompt V1 where we require the model to first present the reasons for the evaluation, and then provide the final score.\\
    
    \bottomrule
    \end{tabular}
    \caption{Prompts designed for Pairwise Comparison and Explicit Score for assessing the quality of storylines in story generation. Note that Prompt V4 of Explicit Score is cited from \cite{wang2023chatgpt}.}
    \label{tab:prompts_pairwise_comparison_story_generation}
\end{table*}

\begin{figure}[!t]
    \centering
    \includegraphics[scale=0.45]{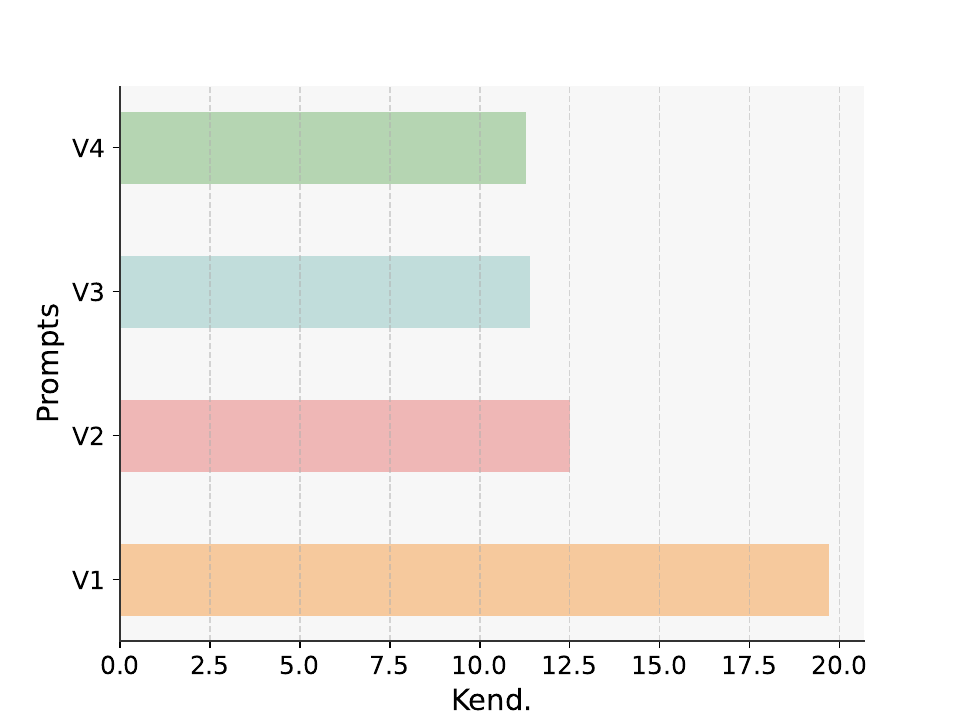}
    \caption{
    Estimated Kendall's tau-b (Kend.) correlation of Pairwise Comparison using ChatGPT with different prompts on OpenMEVA. 
    We use greedy decoding for Prompt V1$\sim$V3. Whereas, for Prompt V4 we use Top-P sampling five times to obtain multiple results and vote for the final decision.}
    \label{fig:pairwise_comparison_across_prompts}
\end{figure}

\begin{figure*}[!htbp]
\centering
\subfigure[{Prompt V1 (Default)}]{
\begin{minipage}[t]{0.33\linewidth}
\centering
\includegraphics[scale=0.42]{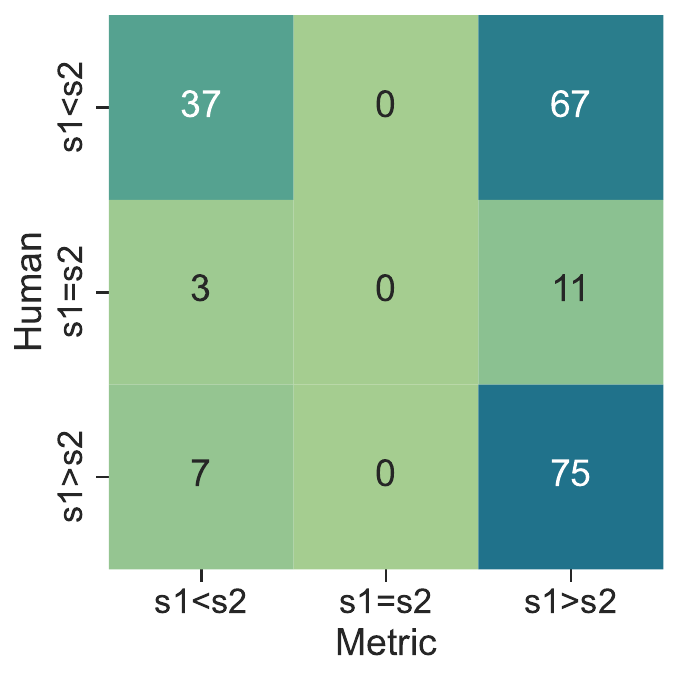}
\label{fig:confusion_matrix_prompt_v1}
\end{minipage}%
}%
\subfigure[Prompt V4 (``Chain-of-Thought'' )]{
\begin{minipage}[t]{0.33\linewidth}
\centering
\includegraphics[scale=0.42]{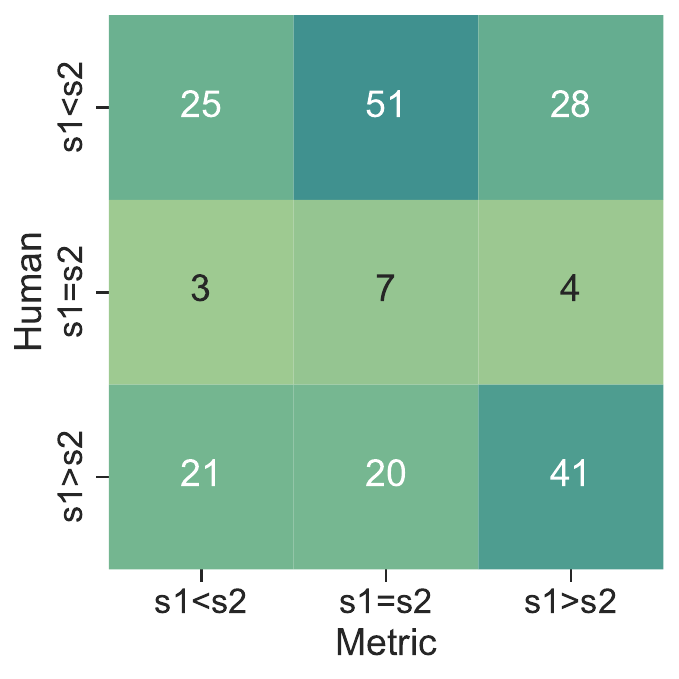}
\label{fig:confusion_matrix_prompt_v4}
\end{minipage}
}%
\subfigure[Prompt V3 (Mirrored)]{
\begin{minipage}[t]{0.33 \linewidth}
\centering
\includegraphics[scale=0.42]{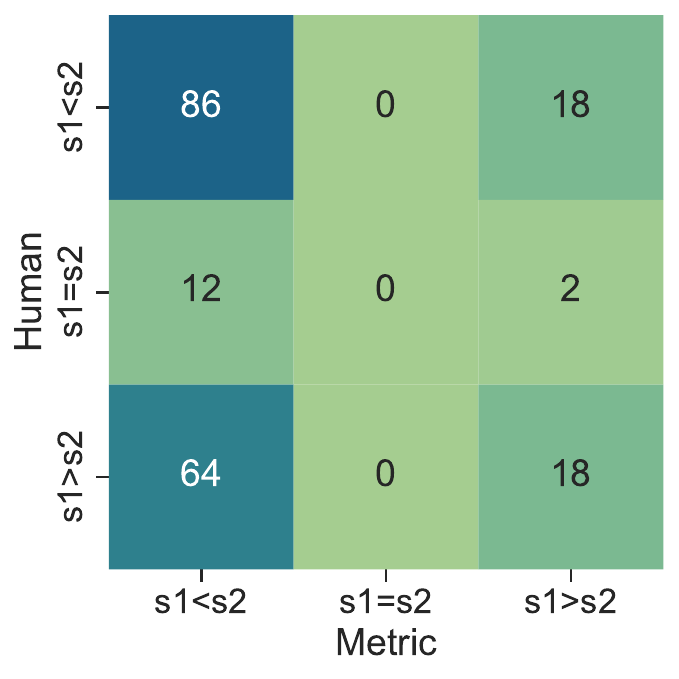}
\label{fig:confusion_matrix_prompt_v3}
\end{minipage}
}%
\centering
\caption{Confusion matrices of pairwise comparisons on OpenMEVA based on different prompts using ChatGPT. 
Prompt V1 is the default prompt used in the main experiments. 
Prompt V4 and V3 are the ``chain-of-thought'' and ``mirrored'' versions of Prompt V1 respectively. 
Details of these prompts are presented in Table~\ref{tab:prompts_pairwise_comparison_story_generation} and Appendix~\ref{sec:different_prompts_pairwise_comparison_story_gen}.}
\label{fig:confusion_matrix}
\end{figure*}

In the main experiments, it is noteworthy that direct pairwise comparison using ChatGPT did not yield satisfactory results. To investigate whether this was caused by poorly designed prompts,  alternative prompts were also evaluated. These prompts are briefly described in Table~\ref{tab:prompts_pairwise_comparison_story_generation}, with detailed information provided in Appendix~\ref{sec:different_prompts_pairwise_comparison_story_gen}. 
Surprisingly, changing the prompt did not improve performance, but rather worsened it, as illustrated in Figure~\ref{fig:pairwise_comparison_across_prompts}.

To gain further insights, we examined the confusion matrices of results based on different prompts, as shown in Figure~\ref{fig:confusion_matrix}. Our analysis revealed that, although we have provided the option of "both storylines equally good" in the default prompt (Prompt V1), ChatGPT still tended to choose one storyline that it deemed "better", as observed from Figure~ \ref{fig:confusion_matrix_prompt_v1}. This could be attributed to the bias introduced by adding "Answer: I will choose Option" at the end of the prompt, which may have induced the model to make a biased choice at the beginning of the answer.
To address this issue, we modified the prompt to require ChatGPT to present its reasoning process before making the final decision (Prompt V4). With this prompt, the model was more likely to choose the "tie" option, as indicated by the ``s1=s2'' column in Figure~\ref{fig:confusion_matrix_prompt_v4}.


After analyzing ChatGPT's reasoning process, we discovered that ChatGPT frequently concludes that "the quality of the two given storylines is equally poor." As a result, we prompted ChatGPT to choose the "worse" storyline instead of the "better" one (Prompt V3). However, this questioning approach did not yield a better outcome. 
In addition, 
Figure~\ref{fig:confusion_matrix_prompt_v3} shows that although Prompt V3 is a mirrored version of Prompt V1, which changes the prompt from selecting the better option to choosing the worse one, ChatGPT's results based on these two prompts are not always consistent. 
For example, in one case, ChatGPT selected Storyline-1 as better based on Prompt V1, but under the guidance of Prompt V3, it may not necessarily choose Storyline-2 as worse. 


Overall, we speculate that the poor quality of the candidate texts used in our experiments is the main reason why comparing pairs directly with ChatGPT did not yield good results. ChatGPT perceives the candidate texts as generally low quality, making it to select a ``better'' or ``worse'' one from them. This might lead to ChatGPT's unstable decisions.


\subsection{Why does Explicit Score generally perform better than Implicit Score?}
\label{sec:analysis_implicit_score}
\begin{figure}[!t]
    \centering
    \includegraphics[scale=0.5]{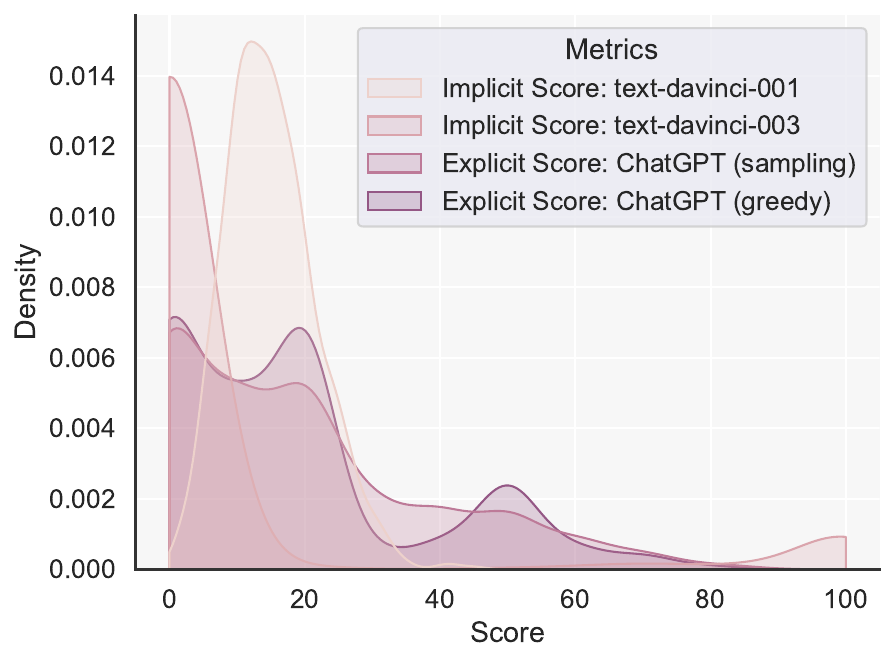}
    \caption{Distribution of different types of Individual Scores on OpenMEVA. The Implicit Score is rescaled into [0,100].}
    \label{fig:score_density}
\end{figure}
\begin{table}[!t]
\small
    \centering
    \begin{tabular}{l c c}
    \toprule
    \textbf{Explicit Score} & \textbf{Spear.} & \textbf{Pear.} \\
    \midrule
    \textsc{ChatGPT} \\
    [0.5ex]\cdashline{1-3}\noalign{\vskip 0.5ex}
    w/ \textsc{Prompt V1 (greedy)} & 49.9 & 51.7 \\
    w/ \textsc{Prompt V2 (greedy)} & \textbf{50.8} & \textbf{53.6} \\
    w/ \textsc{Prompt V3 (greedy)} & 49.4 & 52.0 \\
    w/ \textsc{Prompt V4 (greedy)} & 46.1 & 48.4 \\
    w/ \textsc{Prompt V5 (sampling)} & 47.2 & 50.8 \\
    \bottomrule
    \end{tabular}
    \caption{Sample-level Spearman (Spear.) and Pearson (Pear.) correlation for Explicit Score based on ChatGPT with different prompts on OpenMEVA. We use greedy decoding for Prompt V1$\sim$V4. Whereas, for Prompt V5, we employ Top-P sampling five times to generate multiple reasons and average the resulting scores.}
    \label{tab:ablation_prompts_spearman_openmeva}
\end{table}
\vspace{-0.5mm}
In order to obtain the Explicit Score, we utilize ChatGPT to generate scores in a natural language format. However, as we do not have access to ChatGPT's token probabilities, we instead rely on the confidence of text-davinci series models to determine the Implicit Score, which reflects how well a text meets a particular evaluation criterion. As stated in the Main Experiments (\S \ref{sec:main_experiments}), the Explicit Score is generally more effective than the Implicit Score. This difference in effectiveness could be attributed not only to the variation in the models used but also to the distribution of the two scores. Figure~\ref{fig:score_density} illustrates that the Implicit Score distribution has a peaked structure and is concentrated within a small range. In contrast, the Explicit Score distribution is smoother, allowing for greater discrimination between scores for different texts.


\subsection{How does the prompt design affect Explicit Score?}\label{sec:analysis_explicit_score}

\begin{figure}[!t]
    \centering
    \includegraphics[scale=0.55]{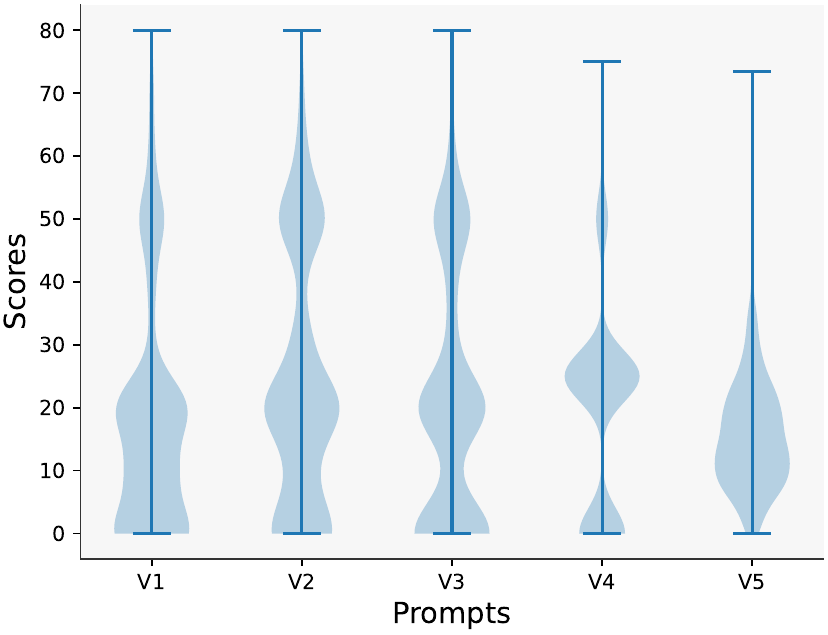}
    \caption{Distribution of Explicit Scores based on ChatGPT with different prompts on OpenMEVA. 
    For Prompt V4, the scores are normalized into [0, 100].}
    \label{fig:explicit_score_distribution_across_prompts}
\end{figure}

We also investigate the impact of prompt design on the performance of rating Explicit Scores generated by ChatGPT. 
The detailed prompts are provided in Appendix~\ref{sec:different_prompts_explicit_score_story_gen}, and their main features and differences are summarized in Table~\ref{tab:prompts_pairwise_comparison_story_generation}. Our results, presented in Table~\ref{tab:ablation_prompts_spearman_openmeva}, indicate that paraphrasing (V2) or simplifying (V3) the default prompt (V1) does not significantly affect the performance of Explicit Score based on ChatGPT. In contrast, refining scoring criteria (V4) or providing reasons before scoring (V5) results in a slight decrease in performance. The former may be due to the fact that the refined scoring rules in Prompt V4 do not fully match the standards used for actual manual annotation, and dividing scores into five scales reduces the distinction between scores for different samples. The latter may be due to the overall low quality of the dataset. Our observation indicates that ChatGPT's evaluations for each text are similar and mostly negative. After giving reasons before scoring, ChatGPT's scoring focuses more on the reasons rather than the text itself, resulting in lower scores for each text based on Prompt V5 and reducing the distinction between scores.
The detailed distribution of scores derived from different prompts is demonstrated using a violin plot in Figure~\ref{fig:explicit_score_distribution_across_prompts}.

\section{Related Work}

In the field of text quality evaluation, researchers have devised two main lines of approaches: reference-based and reference-free methods.
The reference-based text evaluation aims to assess the quality by comparing outputs with ground truth, e.g. ROUGE \cite{lin-2004-rouge}, BERTScore \cite{bert-score} and BARTScore \cite{NEURIPS2021_e4d2b6e6}.
However, due to the inherent complexity and diversity of text, it is impossible to obtain references covering the entire spectrum of potential outputs. 
This limitation has prompted researchers to explore reference-free evaluation methods without relying on predefined references 
e.g. 
{iBLEU} \cite{sun-zhou-2012-joint} and {ParaScore} \cite{shen-etal-2022-evaluation}. 
In this line, a reliable sentence representation model is required \cite{gao2021simcse,shen2023sen2pro,shen2023simple}.
Recent studies have indicated that LLM-based evaluation methods can exhibit good consistency with human evaluation in assessing text quality \cite{fu2023gptscore, wang2023chatgpt, kocmi2023large, ji2023exploring}. 
However, most of these works are preliminary explorations or require gold references. 
On the contrary, we are the first to conduct extensive experiments to investigate the optimal evaluation approaches using LLMs without references, and moreover propose some clues for customized text evaluation.

\section{Conclusion}
This paper explores the feasibility of LLMs, specifically ChatGPT and text-davinci series models, for evaluating text quality in a reference-free mode. Through an empirical study, we compare different methods for the evaluation of text quality and recommend the use of an Explicit Score generated by ChatGPT as the most effective and stable approach. 
This paper also highlights the potential problem of directly comparing the quality of two texts using ChatGPT and the limitations of Implicit Scores obtained through the confidence of text-davinci series models. 
The prompt design is another crucial factor impacting the performance of the Explicit Score generated by ChatGPT. 
Overall, this paper demonstrates the potential of LLMs in evaluating text quality without reference and we hope it will provide useful insights for future research.
\section*{Limitations}


\begin{itemize}[wide=0\parindent]
    \item \textbf{Meta Evaluation Strategy}
    
    We primarily assess the reliability of metrics based on their correlation with human scores. However, it should be noted that the consistency between scores annotated by different raters may not always be high in certain datasets. Hence, the correlation with human ratings may not always reflect the performance of metrics appropriately.

    \item \textbf{Coverage of Texts}
    
    We only conducted experiments on four text-generation tasks. 
    Additionally, the quality distribution of the evaluated texts may be non-uniform, potentially lacking in extremely high-quality texts. 
    Even if a metric performs well in evaluating a set of low-quality texts, it does not necessarily imply the same level of discrimination for high-quality texts, and vice versa. 
    Furthermore, our evaluation has been limited to short texts, omitting the consideration of long-text generation.

    \item \textbf{Coverage of Models}

    We utilize OpenAI's API to access their language models, including ChatGPT (gpt3.5-turbo-0301), text-davinci-003, and text-davinci-001. However, these models may be updated over time, which can result in inconsistencies in experimental outcomes. 
    Moreover, we have not considered a wider range of LLMs, such as text-babbage-001, text-curie-001, and the FLAN-T5 series. 
    Regrettably, due to API limitations, we were unable to obtain results from the more powerful GPT4 model. 

    \item \textbf{Prompt Design}

    
    Our exploration of prompts was limited to a few basic variations. Future research may benefit from more sophisticated prompt designs, such as incorporating few-shot demonstrations, providing more precise annotation guidelines, or guiding the model through multi-turn conversations to facilitate a more accurate assessment of text quality.
\end{itemize}

\section*{Acknowledgements}
This research was supported in part by the National Natural Science Foundation of China(62006062, 62176076), the Guangdong Provincial Key Laboratory of Novel Security Intelligence Technologies(2022B121201000
5), Natural Science Foundation of Guangdong(2023A1515012922), and Key Technologies Research and Development Program of Shenzhen JSGG20210802154400001.

\clearpage
\bibliography{anthology,custom}
\bibliographystyle{acl_natbib}

\clearpage
\appendix
\section{Different Prompts for Explicit Score on Story Generation}
\label{sec:different_prompts_explicit_score_story_gen}
\begin{table}[htb]
\small
    \centering
    \colorbox{gray!8}{
    \begin{tabular}{@{}p{7.3cm}}
    ======= \textsc{Prompt for Explicit Score V1} =======\\\\
    Score the following storyline given the beginning of the story on a continual scale from 0 (worst) to 100 (best), where a score of 0 means "The storyline makes no sense and is totally not understandable" and a score of 100 means "The storyline is perfect-written and highly consistent with the given beginning of the story".\\\\

    The beginning of the story: \\
    \texttt{[Conditioned Text]}\\\\
    
    Storyline: \\
    \texttt{[Generated Text]}\\\\
    
    Score:
    \end{tabular}}
\end{table}

\begin{table}[!htbp]
\small
    \centering
    \colorbox{gray!8}{
    \begin{tabular}{@{}p{7.3cm}}
    ======= \textsc{Prompt for Explicit Score V2} =======\\\\
    On a scale of 0 to 100, evaluate the storyline based on the given beginning. A score of 0 indicates that the storyline is incomprehensible, while a score of 100 means that the storyline is flawlessly written and logically follows from the beginning of the story.\\\\

    The beginning of the story: \\
    \texttt{[Conditioned Text]}\\\\
    
    Storyline: \\
    \texttt{[Generated Text]}\\\\
    
    Score:
    \end{tabular}}
\end{table}

\begin{table}[htb]
\small
    \centering
    \colorbox{gray!8}{
    \begin{tabular}{@{}p{7.3cm}}
    ======= \textsc{Prompt for Explicit Score V3} =======\\\\
    Score the overall quality of the following storyline given the beginning of the story on a continual scale from 0 (worst) to 100 (best). Consider whether the storyline is well-written and consistent with the given beginning of the story.\\\\

    The beginning of the story: \\
    \texttt{[Conditioned Text]}\\\\
    
    Storyline: \\
    \texttt{[Generated Text]}\\\\
    
    Score:
    \end{tabular}}
\end{table}

\begin{table}[htb]
\small
    \centering
    \colorbox{gray!8}{
    \begin{tabular}{@{}p{7.3cm}}
    ======= \textsc{Prompt for Explicit Score V4} =======\\\\
    Score the following storyline given the beginning of the story with one to five stars. Where \\
    \begin{itemize}[wide=0\parindent,noitemsep, topsep=0.5pt]
        \item one star means "Nonsense",

        \item two stars mean "The storyline has some connections with the beginning, but is not understandable",
    
        \item three stars mean "The storyline has some connections with the beginning and is understandable",
    
        \item four stars mean "The storyline is consistent with the beginning and possibly involves a few grammar mistakes",
    
        \item and five stars mean "Perfect storyline and grammar".
    \end{itemize}\\\\
    
    
    
    
    
    The beginning of the story: \\
    \texttt{[Conditioned Text]}\\\\
    
    Storyline: \\
    \texttt{[Generated Text]}\\\\
    
    Stars (1-5):
    \end{tabular}}
\end{table}

\begin{table}[htb]
\small
    \centering
    \colorbox{gray!8}{
    \begin{tabular}{@{}p{7.3cm}}
    ======= \textsc{Prompt for Explicit Score V5} =======\\\\
    Score the following storyline given the beginning of the story on a continual scale from 0 (worst) to 100 (best), where a score of 0 means "The storyline makes no sense and is totally not understandable" and a score of 100 means "The storyline is perfect-written and highly consistent with the given beginning of the story". Please first give your reason carefully (indicated by "Reason:") and then decide your final score (indicated by "Score: 1-100").\\\\

    The beginning of the story: \\
    \texttt{[Conditioned Text]}\\\\
    
    Storyline: \\
    \texttt{[Generated Text]}\\\\
    \end{tabular}}
\end{table}

\section{Different Prompts for Pairwise Comparison on Story Generation}
\label{sec:different_prompts_pairwise_comparison_story_gen}
\begin{table}[!h]
\small
    \centering
    \colorbox{gray!8}{
    \begin{tabular}{@{}p{7.3cm}}
    ===== \textsc{Prompt for Pairwise Comparison V1} =====\\\\

    Consider the following two storylines written according to the given beginning of the story:\\\\

    The beginning of the story:\\
    \texttt{[Conditioned Text]}\\\\
    
    Storyline-1:\\
    \texttt{[Generated Text-1]}\\\\
    
    Storyline-2:\\
    \texttt{[Generated Text-2]}\\\\
    
    Question: Which storyline is better-written and more consistent with the beginning of the story? Please answer with one of the following options.\\\\
    
    Options:\\
    (A) Storyline-1\\
    (B) Storyline-2\\
    (C) Both storylines are equally well-written and consistent with the beginning of the story.\\\\
    
    Answer: I will choose Option
    \end{tabular}}
\end{table}

\begin{table}[!h]
\small
    \centering
    \colorbox{gray!8}{
    \begin{tabular}{@{}p{7.3cm}}
    ===== \textsc{Prompt for Pairwise Comparison V2} =====\\\\

    Question: Which storyline is better-written and more consistent with the beginning of the story? Please answer with one of the following options.\\\\

    The beginning of the story:\\
    \texttt{[Conditioned Text]}\\\\
    
    Options:\\
    (A) \texttt{[Generated Text-1]}\\
    (B) \texttt{[Generated Text-2]}\\
    (C) Both storylines are equally well-written and consistent with the beginning of the story.\\\\
    
    Answer: I will choose Option
    \end{tabular}}
\end{table}

\begin{table}[!h]
\small
    \centering
    \colorbox{gray!8}{
    \begin{tabular}{@{}p{7.3cm}}
    ===== \textsc{Prompt for Pairwise Comparison V3} =====\\\\

    Consider the following two storylines written according to the given beginning of the story:

    The beginning of the story:\\
    \texttt{[Conditioned Text]}\\\\
    
    Storyline-1:\\
    \texttt{[Generated Text-1]}\\\\
    
    Storyline-2:\\
    \texttt{[Generated Text-2]}\\\\
    
    Question: Which storyline has poorer writing and is less consistent with the beginning of the story? Please answer with one of the following options.\\\\
    
    Options:\\
    (A) Storyline-1\\
    (B) Storyline-2\\
    (C) Both storylines are equally poor-written and inconsistent with the beginning of the story.\\\\
    
    Answer: I will choose Option
    \end{tabular}}
\end{table}

\begin{table}[!h]
\small
    \centering
    \colorbox{gray!8}{
    \begin{tabular}{@{}p{7.3cm}}
    ===== \textsc{Prompt for Pairwise Comparison V4} =====\\\\

    Consider the following two storylines written according to the given beginning of the story:

    The beginning of the story:\\
    \texttt{[Conditioned Text]}\\\\
    
    Storyline-1:\\
    \texttt{[Generated Text-1]}\\\\
    
    Storyline-2:\\
    \texttt{[Generated Text-2]}\\\\
    
    Question: Which storyline is better-written and more consistent with the beginning of the story? Please first give your reason carefully (indicated by "Reason:") and then choose one of the following options (indicated by "Answer: A/B/C").\\\\
    
    Options:\\
    (A) Storyline-1\\
    (B) Storyline-2\\
    (C) Both storylines are equally well-written (poor-written) and consistent (inconsistent) with the beginning of the story.
    \end{tabular}}
\end{table}
\clearpage
\section{An Explanation of Kendall's Tall-b}\label{app:kendall}

Kendall's Tau-b is a measure of the correlation between two variables, specifically designed to handle ties and ranks. The formula to calculate Kendall's Tau-b is as follows:
\begin{equation}
\small
    \tau_B = \frac{P-Q}{\sqrt{(P+Q+T)(P+Q+U)}}.
\end{equation}
where P is the number of concordant pairs, Q is the number of discordant pairs, T is the number of ties only in human judgments, and U is the number of ties only in the given metric. 
To better understand the calculation of P, Q, T, and U, we can refer to the following table:
\begin{table}[!htbp]
\small
    \centering
    \begin{tabular}{c c|c c c}
    & & \multicolumn{3}{c}{Metric}\\
    \hline
    & & $s_1 < s_2$ & $s_1 = s_2$ & $s_1 > s_2$  \\
    \multirow{3}{*}{\rotatebox{90}{Human}} & $s_1 < s_2$ & P & U & Q \\
    & $s_1 = s_2$ & T & - & T\\
    & $s_1 > s_2$ & Q & U & P\\
    \end{tabular}
\end{table}
\end{document}